\documentclass{article}
\usepackage{graphicx}
\usepackage[preprint]{corl_2026} 
\usepackage{booktabs}      
\usepackage{multirow}      
\usepackage{amsmath}   
\usepackage{amssymb}   
\usepackage{graphicx}
\usepackage{listings}  
\usepackage{xcolor}

\title{NebulaVLA: A Dual-Frequency Vision-Language-Action Model With Guide Action for Robotic Manipulation}

\author{
  \textbf{Cong Zhao, Shuai Tian, Xu Zhang, Baocheng Ni, Xinguo Song, Xueying Sun, Shu Jiang, Shouchang Yang}\\
  \textbf{Bo Tang, Jin Deng, Ge Zhu, YongCheng Wang$^*$, Jin Xu, Ri Yang}\\
  ZTE Corporation\\
  China\\
  \texttt{wang.yongcheng@zte.com.cn}\\
  $^*$Corresponding author
}

\begin{document}
\maketitle


\begin{abstract}
    Real-world deployment of Vision-Language-Action (VLA) models is often bottlenecked by efficiency-performance trade-offs, cross-embodiment generalization, and execution smoothness. We present NebulaVLA, an asynchronous dual-frequency architecture that decouples high-level semantic reasoning from low-level action control, optimizing computational resources and modularity. To bridge semantic gaps across heterogeneous robots, we introduce GESTURE-7, a unified language-grounded action representation. Furthermore, our Guide Action algorithm enforces kinematic continuity via mask-based smoothness constraints. Comprehensive evaluations demonstrate that NebulaVLA significantly outperforms synchronous baselines, achieving an 85.5\% average success rate on LIBERO-Plus and accelerating action generation by  \textasciitilde 2.7$\times$. This asynchronous design enables highly efficient and responsive control for practical robotics.
\end{abstract}


\section{Introduction}
Robotic manipulation is transitioning from modular pipelines to end-to-end learning, led by Vision-Language-Action (VLA) models. By unifying perception, language, and control within a single architecture, VLAs enable natural language instruction following and complex task execution with enhanced generalization \cite{zitkovich2023rt,driess2023palm}. The field has advanced rapidly: since RT-1 \cite{brohan2022rt} established the Transformer-based policy foundation, OpenVLA \cite{kim2024openvla} accelerated community progress via scalable open-source development, and the InternVLA series \cite{chen2025internvla,cai2026internvlaa1unifyingunderstandinggeneration} introduced precise spatial grounding and a unified understanding-generation-action paradigm.

Contemporary VLA models, however, face three bottlenecks. First, monolithic synchronous architectures enforce a rigid efficiency-performance trade-off: uniform temporal resolution cannot concurrently support low-frequency semantic grounding  and high-frequency motor control. Second, inadequate embodiment representation impedes cross-morphology generalization; mapping continuous kinematics to discrete tokens demands expansive vocabularies and large-scale demonstration data to span the action space, hindering unified action semantics across heterogeneous platforms. Third, maintaining action smoothness and continuity during execution remains challenging. The temporal mismatch between asynchronous real-time inference and action chunk generation often induces pronounced jitter and execution discontinuities.

We introduce \textbf{NebulaVLA}, a dual-frequency VLA model grounded in a fast-slow cognitive paradigm \cite{kahneman2011thinking}, designed to systematically address these limitations.
\begin{enumerate}
    \item \textbf{\textit{Heterogeneous Dual-Frequency Architecture.}} Deliberative reasoning (System 2, 10\,Hz VLM) is decoupled from reactive control (System 1, 20\,Hz DiT generator), optimizing the compute-latency and enabling distributed deployment.
   \item \textbf{\textit{Unified Representation \& Three-Stage Training.}} We propose \textbf{GESTURE-7}, a 7D continuous end-effector vector that unifies action semantics via natural language keyword mapping which enables RL. Training follows a three-stage pipeline: Stage I and II sequentially optimize System 2 via supervised fine-tuning (SFT) and GRPO-based \cite{shao2024deepseekmath} RL to establish robust semantic planning; Stage III performs System 1 SFT using a weighted flow-matching loss to distill priors of S2, ensuring strict training-inference frequency alignment.
    
    \item \textbf{\textit{Guide Action Mechanism.}} Inspired by \textit{Image Outpainting} \cite{pain2021,lugmayr2022repaint}, we condition next-chunk generation on executed trajectory anchors from the preceding chunk; mask constraints during diffusion denoising enforce semantic-guided boundary stitching, significantly improving trajectory continuity and smoothness.
\end{enumerate}

Experiments demonstrate that NebulaVLA significantly reduces inference latency while boosting task success rates over synchronous baselines. GESTURE-7 markedly enhances cross-embodiment transferability, and Guide Action effectively suppresses real-world execution jitter (quantitative results in Sec.~\ref{sec:experiments}). 
\section{Related Work}
\label{sec:related}

\noindent\textbf{\textit{VLA Architectures and Frequency Design.}}
VLA models have evolved from foundational Transformer architectures \cite{brohan2022rt} to large-scale frameworks integrating spatial grounding and unified perception-action paradigms, including knowledge-transfer variants \cite{zitkovich2023rt,driess2023palm}, open-source scalable designs \cite{kim2024openvla}, and unified understanding-generation-action systems such as $\pi$0 series \cite{black2024pi_0,pi2025pi05,intelligence2026pi}, InternVLA series \cite{chen2025internvla,cai2026internvlaa1unifyingunderstandinggeneration} and GR series \cite{wu2024unleashing,cheang2024gr,cheang2025gr}. Despite this progress, these approaches struggle to reconcile the computational overhead of semantic reasoning with the low-latency demands of high-frequency motor control. Recent efforts have begun exploring fast-slow system decoupling: academically, HiRT \cite{zhang2024hirt} conditions high-frequency policies on asynchronously cached VLM features; industrially, Figure Helix \cite{helix2025} adopts a similar dual-frequency hierarchy in real-world deployment. Both validate the efficacy of heterogeneous frequency design in balancing generalization with real-time execution. Aligned with this paradigm, NebulaVLA implements a structured dual-frequency architecture decoupled via explicit data-flow specifications. The design natively supports distributed deployment, minimizing interaction latency through asynchronous communication and workload separation, thereby establishing a scalable structural foundation for synchronizing high-level reasoning with low-level control.

\noindent\textbf{\textit{Embodiment Representation and Action Tokenization.}}
Cross-embodiment generalization fundamentally relies on unified action representation. Current discrete tokenization methods (FAST \cite{pertsch2025fast}, Mind-to-Hand \cite{tang2025mind}) map continuous kinematics to categorical tokens via frequency bands or trajectory clustering, demanding extensive demonstrations and struggling with cross-morphology semantic alignment. Conversely, continuous approaches (PUMA/DOMINO \cite{fang2026pumadomino}; KineVLA \cite{han2026kinevla}) utilize optical flow or kinesthetic encoding but regress raw states without cross-platform semantic abstraction. NebulaVLA addresses this via \textbf{GESTURE-7}, a hybrid 7D end-effector representation that maps synonymous actions to shared natural language keywords. This formulation natively unifies heterogeneous embodiments (e.g., dexterous hands, grippers, arms) and seamlessly integrates with language-conditioned RL.

\noindent\textbf{\textit{Trajectory Smoothness and Chunk Boundary Continuity.}}
Ensuring motion continuity during asynchronous real-world deployment remains a critical bottleneck. While chunk-based generation (e.g., ACT \cite{zhao2023act}) simplifies policy learning, temporal misalignment between inference latency and control frequency induces severe jitter and boundary discontinuities. Existing solutions predominantly rely on \textit{post-hoc} smoothing: spline-based interpolation \cite{yang2026abpolicy,huang2025notvla,zhao2025vla}, jerk-minimizing RL rewards \cite{li2026smoothvla}, or detokenizer redesigns. Although effective in mitigating artifacts, these methods operate outside the generative process and cannot guarantee semantic consistency across chunks. In contrast, NebulaVLA internalizes smoothness constraints via the \textbf{Guide Action} mechanism. Drawing inspiration from \textit{Image Outpainting} \cite{pain2021,lugmayr2022repaint}, we condition next-chunk generation on executed trajectory anchors and enforce mask constraints during diffusion denoising. This preserves known actions at each step, ensures semantic continuity at overlapping boundaries, and removes the need for post-processing smoothing.
\section{Method}
\label{sec:method}

\subsection{Architecture Overview}
\begin{figure}[htbp]
    \centering
    \includegraphics[width=0.8\textwidth]{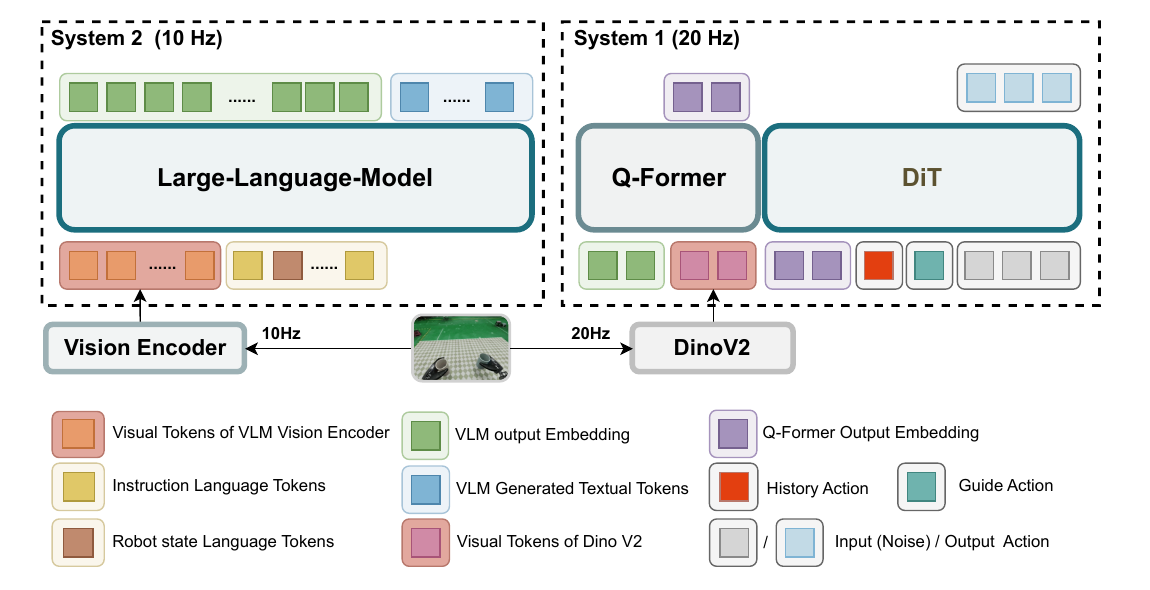}
    \caption{NebulaVLA Architecture. System 2 uses Qwen3-VL as backbone to output abstract action guidance from user instructions and sparse visual inputs. System 1 utilizes a Q-Former to compress S2's guidance alongside high-frequency visual inputs, and ultimately outputs execution actions through a denoising process with a Diffusion Transformer.}
    \label{fig:arch}
\end{figure}

As illustrated in Fig.\ref{fig:arch}, NebulaVLA employs a hierarchical fast-slow architecture with heterogeneous frequencies. System 2 performs high-level semantic planning by processing natural language instructions and sparse visual inputs (\textasciitilde 10\,Hz) to output abstract action guidance. System 1 operates as a low-level controller (\textasciitilde 20\,Hz), conditioning on this guidance with high-frequency visual inputs and action history. By explicitly decoupling these frequencies, our design balances computational efficiency with control responsiveness, natively supporting distributed deployment that minimizes latency in real-world interactions and maximizes hardware throughput.

\subsection{NebulaVLA Components}
\textit{\textbf{System 2 (S2)}}. S2 employs Qwen3-VL \cite{bai2025qwen3} as its vision-language backbone. It extracts hidden-layer features via learnable tokens to serve as abstract action guidance for S1.

\textbf{\textit{System 1 (S1)}}. Inspired by InternVLA-M1 \cite{chen2025internvla}, the design of S1 introduces two further optimizations: (1) contrastive fine-tuning of the pre-trained DINOv2 \cite{oquab2023dinov2} using visual inputs and end-effector states to improve visual-proprioceptive alignment; and (2) integrating the Guide Action mechanism to ensure trajectory continuity and smoothness.

\begin{figure}[htbp]
    \centering
    \includegraphics[width=0.7\textwidth]{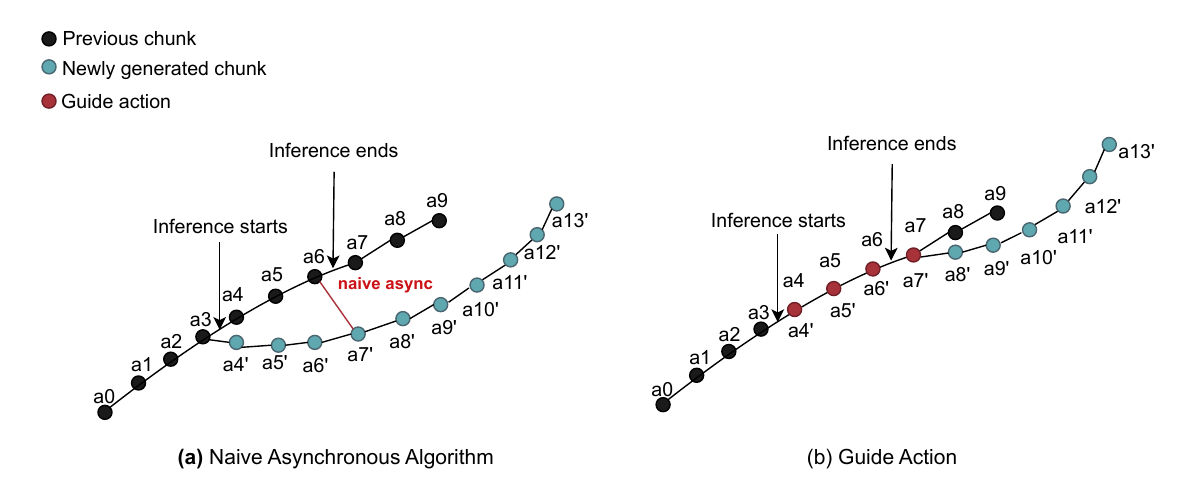}
    \caption{Scheme of the Guide Action mechanism. (a) The naive asynchronous baseline performs a hard switch (red line) after inference ends, leading to execution jitter. (b) Guide Action utilizes unexecuted prior steps (red dots) to condition the generation of the subsequent chunk, ensuring seamless trajectory alignment.}
    \label{fig:guide}
\end{figure}

\textbf{\textit{Guide Action}}. Asynchronous policy inference and action execution in real-world settings often compromise trajectory smoothness. To ensure seamless transitions between action chunks, we propose Guide Action. Inspired by \textit{Image Outpainting} \cite{pain2021,lugmayr2022repaint}, it embeds chunk concatenation directly into the generative pipeline, bypassing conventional RTC \cite{black2026rtc} and other post-hoc smoothing methods \cite{zhao2023act}. As shown in Fig.\ref{fig:guide}a, a hard switch (red line) of trajectory for baseline methods causes mechanical jitter. Instead, we condition the next chunk's generation on prior unexecuted actions (e.g., $a_4$–$a_7$ in Fig.\ref{fig:guide}b). This enforces trajectory alignment ($a'_4$–$a'_7$) for seamless boundary transitions. With guide action points number adapting to inference latency (randomly sampled from 0 to 3 during training), the denoising process is reformulated as the following conditional probabilistic model: 
\begin{equation}
    p_\theta(A_{full}^{t-1} \mid A_{full}^t, A_{prior}, \tau)
\end{equation}
where:
\begin{itemize}
    \item $A_{full}^t$ is the complete action chunk at denoising timestep $t$ (length $T$).
    \item $A_{prior}$ denotes the prior action set, consisting of the executed history action (optional) and the guide action to be executed.
    \item $\tau$ represents the temporal positional query (positional encoding vector).
\end{itemize}

After each denoising step, similar to how \textit{Image Outpainting} preserves original pixels, we forcibly reinstate the known actions:
\begin{equation}
    A_{full}^{t-1} = M \odot A_{prior}^{(t-1)} + (1 - M) \odot A_{noisy}^{t-1}
\end{equation}
where:
\begin{itemize}
    \item $M$ is a binary mask, $M \in \{0, 1\}^T$, such that:
    \begin{itemize}
        \item $M[1..k] = 1$ (anchor region, strictly preserved),
        \item $M[k+1..T] = 0$ (generation region, predicted by the model);
    \end{itemize}
    \item $A_{prior}^{(t-1)}$ is the noisy version of the known actions at diffusion timestep $t-1$;
    \item $A_{noisy}^{t-1}$ is the current denoised result predicted by the model.
\end{itemize}
and $\odot$ denotes element-wise multiplication.

\subsection{Training Process}
NebulaVLA's training pipeline comprises three stages: Stage 1 and Stage 2 perform supervised fine-tuning (SFT) and reinforcement learning (RL) on System 2 based on GESTURE-7 action description, respectively, while Stage 3 executes supervised fine-tuning on System 1.

GESTURE-7, short for Gesture Enumerated State Token for Unified Robot End-effector-7D, is a natural language-based method for describing end-effector states. Specifically, the current state of an end-effector is encoded as a 7-dimensional vector: $[x, y, z, r, p, w, \textit{gesture}]$. In this vector, $x$, $y$, $z$ represent the Cartesian position of the end-effector, while $r$, $p$, $w$ correspond to its orientation (roll, pitch, yaw). The \textit{gesture} component uses a natural language keyword to indicate the target hand posture. Different embodiments performing the same action share the same keyword—for example, the grasping action, whether executed with a parallel gripper or a dexterous hand, is uniformly represented as \textit{grasp}. Thus, GESTURE-7 unifies action descriptions across heterogeneous robots. More details are provided in Appendix \ref{sec:AppendixA}.

\textbf{\textit{Stage 1: System 2 Supervised Fine-Tuning.}} Stage 1 establishes System 2's visual-grounded proprioceptive understanding and trajectory prediction capabilities through supervised fine-tuning. The training mixture is organized into three task groups: (1) spatial grounding, including end-effector pose regression in GESTURE-7 format and 2D keypoint regression from visual observations; (2) trajectory forecasting, encompassing both 2D waypoint and 3D trajectory point prediction at 2 Hz; and (3) reasoning and planning, where the model generates chain-of-thought subtask decompositions paired with future GESTURE-7 trajectories, conditioned on visual inputs, language instructions, and current end-effector states. Details are provided in Appendix \ref{sec:AppendixB}.

\textbf{\textit{Stage 2: System 2 Reinforcement Learning.}} Reinforcement learning has proven highly effective for aligning large language and vision-language models. Following Stage 1, we further enhance System 2's trajectory planning capability through GRPO-based \cite{shao2024deepseekmath} reinforcement learning, retaining the same task suite as in supervised fine-tuning. The reward function for each task is detailed in Appendix \ref{sec:AppendixC}.

\textbf{\textit{Stage 3: Supervised Fine-Tuning with Continuous Action.}} Stage 3 jointly optimizes the weights of both System 2 and System 1, while keeping their respective visual encoders frozen throughout training. We adopt an L2 loss to supervise the continuous action outputs. Owing to the guided action mechanism, the loss is restricted to the newly generated action region where the binary mask $M=0$ (i.e., time steps $[k+1..T]$ following the provided guide actions of length $k$):
\begin{equation}
\mathcal{L}(\theta) = E_{{\epsilon},t} \Bigl[ 
  \bigl\| ({1} - {M}) \odot ({\epsilon} - {\epsilon}_\theta({x}_t, t)) \bigr\|^2 
\Bigr] \label{eq:mask_loss}
\end{equation}

Here, $\epsilon$ denotes the ground-truth noise, $\epsilon_\theta(\cdot)$ is the noise prediction from the model parameterized by $\theta$, $M \in \{0,1\}$ is a binary mask with $M=1$ indicating the guide action context (positions $[1..k]$) and $M=0$ indicating the newly generated region.

\section{Experiment}
\label{sec:experiments}
\subsection{Experiment Setup}

\textbf{\textit{Simulation Experiments.}} We employ LIBERO-Plus \cite{fei25libero-plus}, a large-scale embodied manipulation benchmark designed to rigorously assess vision-language-action (VLA) policies under controlled environmental perturbations. Compared to the LIBERO \cite{liu2023libero} framework, LIBERO-Plus introduces 7 perturbation dimensions (encompassing 21 sub-dimensions), across 10,030 programmatically generated tasks categorized into five difficulty levels (L1–L5). we adopt the Success Rate as the primary performance metric. This includes success rates under individual and combined perturbation dimensions, as well as task-wise success rates.

\textbf{\textit{Real-World Experiments.}} We conduct experiments to validate real-world deployment performance—particularly the effectiveness of the heterogeneous-frequency architecture and the Guide Action mechanism on AgiBot A2 robot platform, which features dual-arm manipulation capabilities suitable for complex industrial and domestic service scenarios. We select two representative manipulation tasks: Pick-and-Place and Packaging Line Material Feeding, corresponding to general grasping scenarios and structured industrial scenarios, respectively. Training is conducted on a single node with 8$\times$H800 GPUs using a batch size of 96.

We employ Success Rate (SR) and Inference Latency to evaluate task completion capability and computational efficiency, and use Joint Jerk to measure trajectory smoothness. For each joint $j$, the mean absolute jerk is computed as:
\begin{equation}
  \bar{J}_j = \frac{1}{N} \sum_{i=1}^{N} \left| \frac{q_{i+1,j} - 3q_{i,j} + 3q_{i-1,j} - q_{i-2,j}}{\Delta t^3} \right|
  \label{eq:jerk}
\end{equation}

  where $q_{i,j}$ denotes the angle of the $j$-th joint at the $i$-th action step along the trajectory. $\Delta t$ is the control period, and $N$ is the total number of trajectory points. The metric $\bar{J}_j$ is reported independently for each of the 7 joints in the right arm.

\textbf{\textit{Baseline Methods.}} We compare NebulaVLA against state-of-the-art open-source VLA systems including InternVLA-M1 \cite{chen2025internvla}, ${\pi0.5}$ \cite{pi2025pi05}, GR00T N1.5 \cite{bjorck2025gr00t}, and OpenVLA-OFT \cite{kim2025fine}. We directly adopt the officially reported LIBERO-Plus results for those that have reported them. For methods lacking official results, we reproduce them using publicly released checkpoints and the training pipeline from LeRobot \cite{cadene2024lerobot} library to ensure identical training configurations. All methods are evaluated under identical LIBERO-Plus observation spaces, action spaces, and evaluation protocols for a fair comparison.

\subsection{Main Results}
\begin{figure}[htbp]
    \centering
    \includegraphics[width=0.8\textwidth]{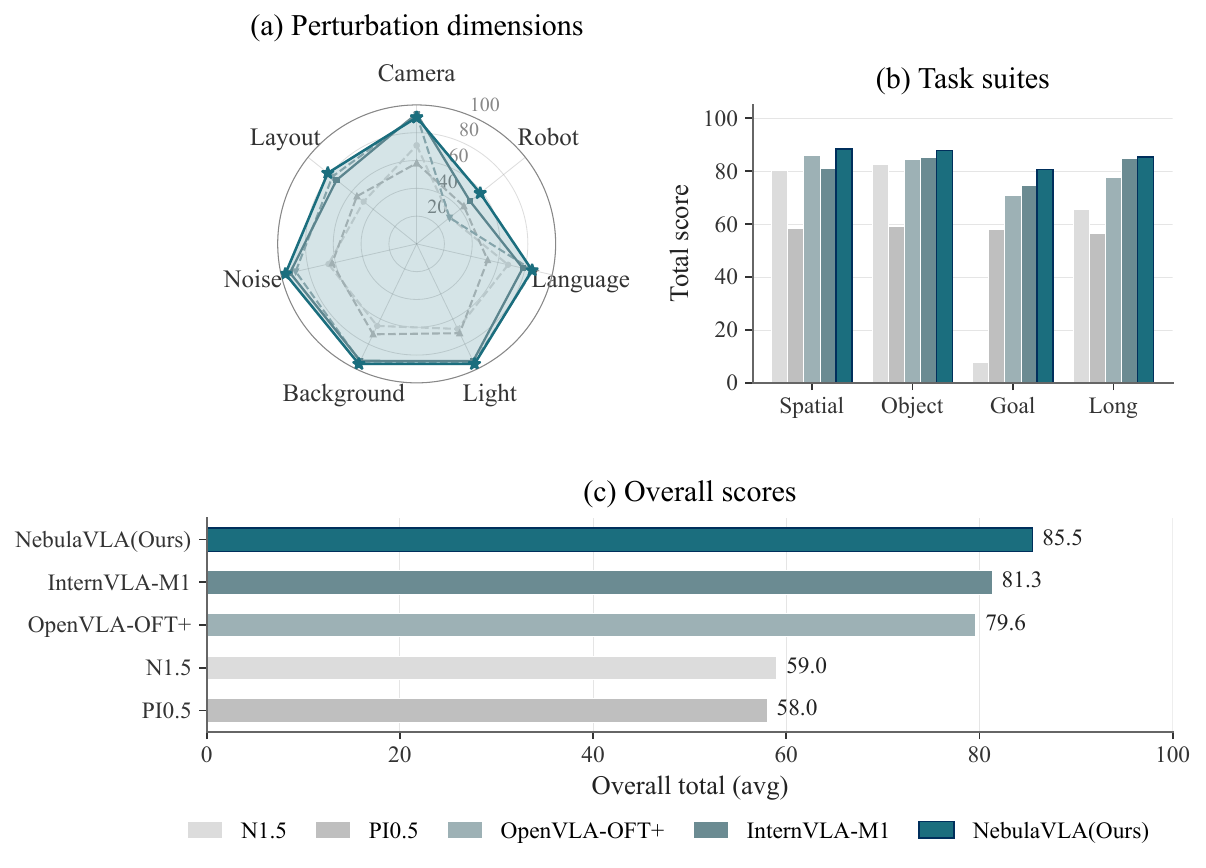}
    \caption{Simulation Result Comparison on Libero-Plus}
    \label{fig:example}
\end{figure}

\textbf{\textit{Simulation Results.}} As shown in Fig. \ref{fig:example}, NebulaVLA achieves state-of-the-art robustness on LIBERO-Plus with an overall success rate of 85.5\%, surpassing InternVLA-M1 (81.3\%) by 4.2\% and outperforming OpenVLA-OFT+ (79.6\%), GR00T N1.5 (59.0\%), and $\pi$0.5 (58.0\%).

NebulaVLA maintains near-optimal performance across dimensions: near-perfect Language scores, marked improvements over InternVLA-M1 and OpenVLA-OFT+ in Robot Initial State, $\sim$8\% gains in Layout, and comparable or superior results in Light, Background, and Noise. It further leads in Spatial, Object, and Goal (6\% and 10\% over InternVLA-M1 and OpenVLA-OFT+, respectively), matches InternVLA-M1 in Long while outperforming OpenVLA-OFT+ by $\sim$8\%, and maintains 2--7\% margins in Spatial and Object.

Notably, NebulaVLA exhibits marginally lower Camera performance than InternVLA-M1, attributed to GESTURE-7's training hypothesis establishing fixed visual-spatial mappings (see ablation experiments).

\textbf{\textit{Real-World Results.}} Real-world experiments are conducted on the AgiBot A2 robot platform with stereo RGB camera inputs. The policy is deployed on a NVIDIA RTX 4070 GPU. We select Task 1 (Pick-and-Place) and Task 2 (Packaging Line Material Feeding) for evaluation, comparing against InternVLA-M1. As shown in Table \ref{tab:comparison}, NebulaVLA-Heterogeneous achieves the highest success rates in both tasks while maintaining the lowest inference latency.
\begin{table}[htbp]
    \centering
    \begin{tabular*}{\columnwidth}{@{\extracolsep{\fill}} lccc}
        \toprule
        & \textbf{InternVLA-M1} & \textbf{NebulaVLA-Homo} & \textbf{NebulaVLA-Heter} \\
        \midrule
        Task 1: Pick-and-Place SR & 77.91\%& 87.14\% & 92.08\% \\
        Task 2: Feeding SR       & 72.5\%  & 90\%    & 92.5\%  \\
        Avg. Inference Latency (ms)    & 81      & 115     & 42      \\
        \bottomrule
    \end{tabular*}
    \caption{Real-World Performance Comparison with InternVLA-M1}
    \label{tab:comparison}
\end{table}

The heterogeneous-frequency structure significantly outperforms both the homogeneous-frequency variant and InternVLA-M1 in terms of success rate and latency. Compared to the homogeneous-frequency design, the heterogeneous architecture reduces the average step latency from 115 ms to 42 ms while improving success rates. This advantage stems from the decoupling of System 1 (fast action response) and System 2 (deep semantic reasoning): the high-frequency path focuses on lightweight action policy execution, while the low-frequency path handles scene understanding and task planning at longer intervals, thereby balancing efficiency and performance.

Furthermore, the heterogeneous-frequency architecture offers significant engineering advantages for real-world deployment. Since System 1 and System 2 operate independently, they can be deployed on separate computational devices. This enables NebulaVLA to flexibly adapt to edge computing scenarios, providing a practical solution for efficient VLA deployment in real robotic systems.

\subsection{Ablation Studies}
\textbf{\textit{Training Strategy Ablation Study.}} To validate the effectiveness of our three-stage training strategy, we conduct ablation experiments across the seven perturbation dimensions (Camera, Robot, Language, Light, Background, Noise, Layout) on the LIBERO-Plus benchmark, with results presented in Table \ref{tab:libero-plus-ablation}. NebulaVLA's Stage 1 and Stage 2 perform S2 supervised fine-tuning (SFT) and reinforcement learning (RL), respectively, based on GESTURE-7, while Stage 3 executes S1 supervised fine-tuning. We configure three comparison settings: (1) NebulaVLA(w/o SFT, w/o RL): the S2 module directly loads open-source pretrained weights, omitting both GESTURE-7-based SFT and RL; (2) NebulaVLA(w/o RL): trained with GESTURE-7-based S2 SFT but omitting S2 RL; (3) NebulaVLA(ALL): complete three-stage trained.

\textbf{\textit{Quantitative Analysis.}} NebulaVLA(ALL) achieves an overall success rate of 85.5\%, representing 1.7\% improvement over NebulaVLA(w/o SFT, w/o RL) (83.8\%), and 2.5\% improvement over NebulaVLA(w/o RL) (83.0\%). This validates the effectiveness of progressive stage-wise optimization. Specifically, S2 SFT significantly improves performance in the Layout dimension (+6.2\%), establishing reliable spatial priors for the policy; S2 RL further enhances performance in semantically-coupled spatial dimensions such as Language and Layout beyond the SFT-only baseline.

Interestingly, we observe performance degradation in the Camera perturbation dimension from both S2 SFT and RL. We attribute this to the influence of GESTURE-7's training hypothesis, where its behavior representation implicitly establishes fixed mappings between visual features and spatial coordinates during training. Camera pose perturbations disrupt this correspondence, leading to degraded performance in this dimension. In future work,  we plan to incorporate observation viewpoint conditioning into the GESTURE-7 training framework, explicitly modeling the mapping between viewpoint variations and spatial coordinates to further improve robustness under camera perturbations.
\begin{table}[htbp]
  \centering
  \small 
  \begin{tabular*}{\columnwidth}{@{\extracolsep{\fill}} lcccccccc}
    \toprule
    & \textbf{Camera} & \textbf{Robot} & \textbf{Lang.} & \textbf{Light} & \textbf{Bkg.} & \textbf{Noise} & \textbf{Layout} & \textbf{Total} \\
    \midrule
    w/o SFT, w/o RL & 93.5 & 57.9 & 80.5 & 97.8 & 94.9 & 92.6 & 73.0 & 83.8 \\
    w/o RL          & 87.8 & 56.3 & 82.8 & 93.3 & 93.2 & 94.2 & 79.2 & 83.0 \\
    NebulaVLA(ALL)  & 91.0 & 58.5 & 85.2 & 95.9 & 95.9 & 96.9 & 81.8 & 85.5 \\
    \bottomrule
  \end{tabular*}
  \caption{Ablation Results of NebulaVLA on LIBERO-Plus}
  \label{tab:libero-plus-ablation}
\end{table}

\begin{figure}[htbp]
    \centering
    \includegraphics[width=0.6\textwidth]{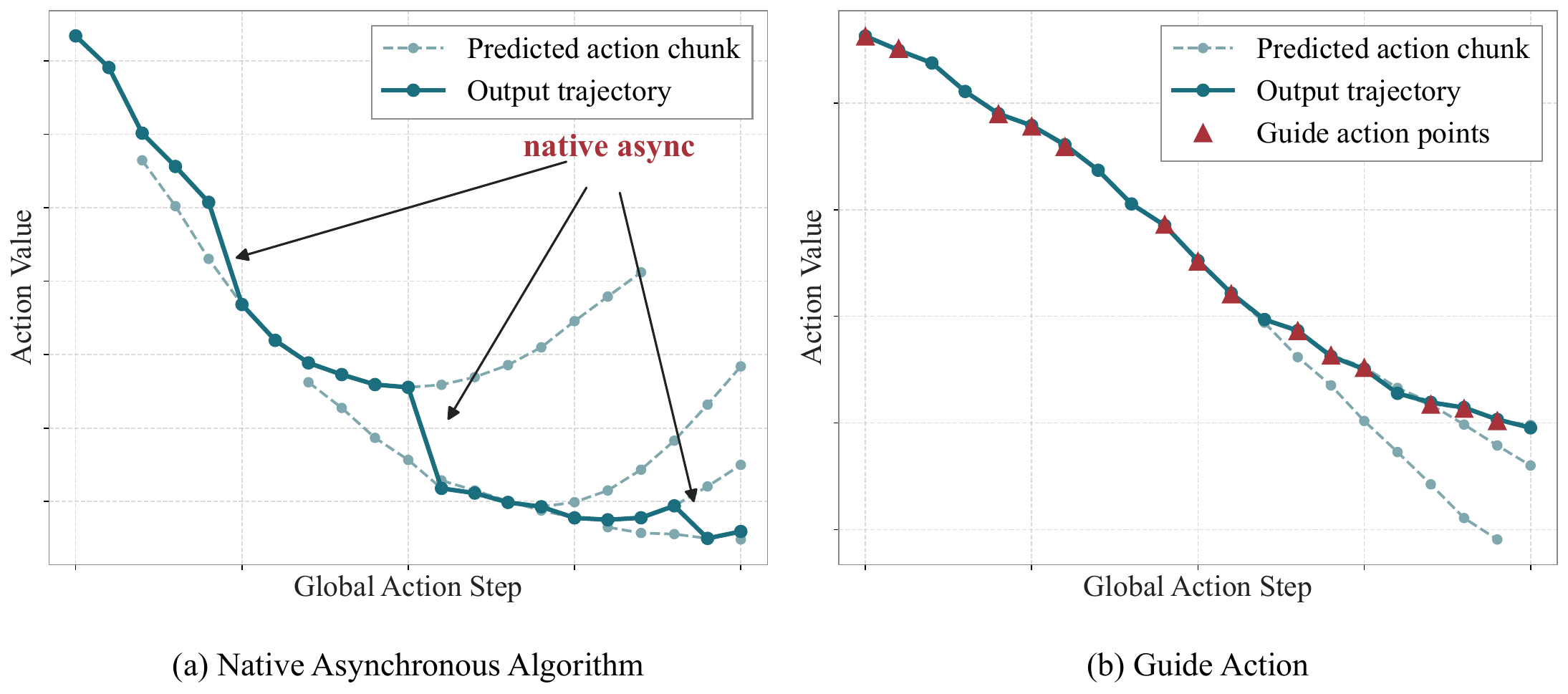}
    \caption{Visualization of \textit{Guide Action} Effectiveness}
    \label{fig:joint-traj}
\end{figure}

\textbf{\textit{Guide Action Ablation Study}} As illustrated in Fig. \ref{fig:joint-traj}, joint trajectory visualization reveals pronounced jumps at action chunk boundaries under the native asynchronous algorithm (left), where means actual outputs deviate from next predicted chunks (arrows), causing abrupt joint variations. The Guide Action algorithm (right) eliminates these discontinuities by introducing preceding guide points (red triangles) that align each new chunk's starting state with the previous trajectory, ensuring smooth joint transitions.

\textbf{\textit{Quantitative Analysis.}} Table \ref{tab:joint-comparison} reports mean jerk ($\times 10^{-3}$) across joints 7–13 over 10 real-world trials. Guide Action reduces jerk in all joints, most substantially in joints 12 and 13 by 41.2\% (4.95→2.91) and 45.1\% (3.37→1.85), respectively, achieving an average reduction of 25.6\% over the native asynchronous baseline. This validates that the Guide Action algorithm effectively suppresses boundary discontinuities and maintains kinematic continuity during chunk transitions.
\begin{table}[htbp]
    \centering
    \begin{tabular}{lccccccc}
        \toprule
        \textbf{Algorithm} & \textbf{Joint 7} & \textbf{Joint 8} & \textbf{Joint 9} & \textbf{Joint 10} & \textbf{Joint 11} & \textbf{Joint 12} & \textbf{Joint 13} \\
        \midrule
        Native Async & 6.18 & 6.47 & 6.55 & 7.66 & 9.69 & 4.95 & 3.37 \\
        Guide Action & 5.27 & 5.08 & 5.47 & 5.59 & 8.43 & 2.91 & 1.85 \\
        \bottomrule
    \end{tabular}
    \caption{Effect of \textit{Guide Action} on Trajectory Smoothness}
    \label{tab:joint-comparison}
\end{table}
\section{Conclusions and Limitations}
\label{sec:conclusion}
\textbf{Conclusions} This paper presents \textbf{NebulaVLA}, a dual-frequency Vision-Language-Action model that systematically addresses three core challenges in robotic manipulation: computational efficiency versus control fidelity, cross-embodiment action representation, and trajectory continuity in real-world deployment. Our key contributions are: (1) a heterogeneous dual-frequency architecture that decouples semantic planning (\textasciitilde 10\,Hz) from reactive motor control (\textasciitilde 20\,Hz), enabling modular deployment and explicit compute-latency optimization; (2) \textbf{GESTURE-7}, a 7D continuous end-effector representation that unifies action semantics across heterogeneous platforms via natural language keyword mapping and natively supports reinforcement learning; and (3) the \textbf{Guide Action} mechanism, which enforces smoothness constraints intrinsically during diffusion denoising to suppress boundary jitter without post-processing.

\textbf{Limitations} Experimental results demonstrate that NebulaVLA consistently outperforms synchronous baselines across diverse manipulation tasks in both simulation and real-world deployment, validating the efficacy of heterogeneous frequency coordination. Nevertheless, several limitations warrant further investigation. First, the current frequency allocation is empirically determined and static; optimal configurations may vary across task domains, and dynamic adaptive scheduling remains an open challenge. Second, while Guide Action effectively suppresses inter-chunk discontinuities, its reliance on fixed guide anchor points introduces minor response latency in rapidly changing environments. Third, practical deployment still faces engineering hurdles in heterogeneous hardware synchronization, multi-sensor timestamp alignment, and distributed clock consistency.
\clearpage
\acknowledgments{This work was supported by the National Key Research and Development Program of China under Grant No. 2025YFB4712700, for the project "Research and Application Demonstration of Trustworthy Lightweight Large Models for Dual-Arm Mobile Robots."}
\bibliography{example}  

@article{brohan2022rt,
  title={Rt-1: Robotics transformer for real-world control at scale},
  author={Brohan, Anthony and Brown, Noah and Carbajal, Justice and Chebotar, Yevgen and Dabis, Joseph and Finn, Chelsea and Gopalakrishnan, Keerthana and Hausman, Karol and Herzog, Alex and Hsu, Jasmine and others},
  journal={arXiv preprint arXiv:2212.06817},
  year={2022}
}

@inproceedings{zitkovich2023rt,
  title={Rt-2: Vision-language-action models transfer web knowledge to robotic control},
  author={Zitkovich, Brianna and Yu, Tianhe and Xu, Sichun and Xu, Peng and Xiao, Ted and Xia, Fei and Wu, Jialin and Wohlhart, Paul and Welker, Stefan and Wahid, Ayzaan and others},
  booktitle={Conference on Robot Learning},
  pages={2165--2183},
  year={2023},
  organization={PMLR}
}

@article{driess2023palm,
  title={Palm-e: An embodied multimodal language model},
  author={Driess, Danny and Xia, Fei and Sajjadi, Mehdi SM and Lynch, Corey and Chowdhery, Aakanksha and Ichter, Brian and Wahid, Ayzaan and Tompson, Jonathan and Vuong, Quan and Yu, Tianhe and others},
  journal={arXiv preprint arXiv:2303.03378},
  year={2023}
  }

@article{kim2024openvla,
  title={Openvla: An open-source vision-language-action model},
  author={Kim, Moo Jin and Pertsch, Karl and Karamcheti, Siddharth and Xiao, Ted and Balakrishna, Ashwin and Nair, Suraj and Rafailov, Rafael and Foster, Ethan and Lam, Grace and Sanketi, Pannag and others},
  journal={arXiv preprint arXiv:2406.09246},
  year={2024}
}

@article{chen2025internvla,
  title={Internvla-m1: A spatially guided vision-language-action framework for generalist robot policy},
  author={Chen, Xinyi and Chen, Yilun and Fu, Yanwei and Gao, Ning and Jia, Jiaya and Jin, Weiyang and Li, Hao and Mu, Yao and Pang, Jiangmiao and Qiao, Yu and others},
  journal={arXiv preprint arXiv:2510.13778},
  year={2025}
}

@misc{cai2026internvlaa1unifyingunderstandinggeneration,
      title={InternVLA-A1: Unifying Understanding, Generation and Action for Robotic Manipulation}, 
      author={Junhao Cai and Zetao Cai and Jiafei Cao and Yilun Chen and Zeyu He and Lei Jiang and Hang Li and Hengjie Li and other},
      year={2026},
      eprint={2601.02456},
      archivePrefix={arXiv},
      primaryClass={cs.RO},
      url={https://arxiv.org/abs/2601.02456}, 
}

@misc{helix2025,
  author       = {{Figure AI}},
  title        = {Helix: A Vision-Language-Action Model for Generalist Humanoid Control},
  howpublished = {Online},
  year         = {2026},
  url          = {https://www.figure.ai/news/helix},
}

@inproceedings{wu2024unleashing,
  title={Unleashing large-scale video generative pre-training for visual robot manipulation},
  author={Wu, Hongtao and Jing, Ya and Cheang, Chilam and Chen, Guangzeng and Xu, Jiafeng and Li, Xinghang and Liu, Minghuan and Li, Hang and Kong, Tao},
  booktitle={International Conference on Learning Representations},
  volume={2024},
  pages={10641--10662},
  year={2024}
}

@article{cheang2024gr,
  title={Gr-2: A generative video-language-action model with web-scale knowledge for robot manipulation},
  author={Cheang, Chi-Lam and Chen, Guangzeng and Jing, Ya and Kong, Tao and Li, Hang and Li, Yifeng and Liu, Yuxiao and Wu, Hongtao and Xu, Jiafeng and Yang, Yichu and others},
  journal={arXiv preprint arXiv:2410.06158},
  year={2024}
}

@article{cheang2025gr,
  title={Gr-3 technical report},
  author={Cheang, Chilam and Chen, Sijin and Cui, Zhongren and Hu, Yingdong and Huang, Liqun and Kong, Tao and Li, Hang and Li, Yifeng and Liu, Yuxiao and Ma, Xiao and others},
  journal={arXiv preprint arXiv:2507.15493},
  year={2025}
}

@article{zhang2024hirt,
  title={Hirt: Enhancing robotic control with hierarchical robot transformers},
  author={Zhang, Jianke and Guo, Yanjiang and Chen, Xiaoyu and Wang, Yen-Jen and Hu, Yucheng and Shi, Chengming and Chen, Jianyu},
  journal={arXiv preprint arXiv:2410.05273},
  year={2024}
}

@article{black2024pi_0,
  title={$\pi_0$: A Vision-Language-Action Flow Model for General Robot Control},
  author={Black, Kevin and Brown, Noah and Driess, Danny and Esmail, Adnan and Equi, Michael and Finn, Chelsea and Fusai, Niccolo and Groom, Lachy and Hausman, Karol and Ichter, Brian and others},
  journal={arXiv preprint arXiv:2410.24164},
  year={2024}
}

@article{bjorck2025gr00t,
  title={Gr00t n1: An open foundation model for generalist humanoid robots},
  author={Bjorck, Johan and Casta{\~n}eda, Fernando and Cherniadev, Nikita and Da, Xingye and Ding, Runyu and Fan, Linxi and Fang, Yu and Fox, Dieter and Hu, Fengyuan and Huang, Spencer and others},
  journal={arXiv preprint arXiv:2503.14734},
  year={2025}
}

@article{han2026kinevla,
  title={KineVLA: Towards Kinematics-Aware Vision-Language-Action Models with Bi-Level Action Decomposition},
  author={Han, Gaoge and Gao, Zhengqing and Li, Ziwen and Huang, Jiaxin and Huang, Shaoli and Karray, Fakhri and Gong, Mingming and Liu, Tongliang},
  journal={arXiv preprint arXiv:2603.17524},
  year={2026}
}

@article{fang2026pumadomino,
  title={Towards Generalizable Robotic Manipulation in Dynamic Environments},
  author={Fang, Heng and Li, Shangru and Wang, Shuhan and Xi, Xuanyang and Liang, Dingkang and Bai, Xiang},
  journal={arXiv preprint arXiv:2603.15620},
  year={2026}
}

@article{zhao2023act,
  title={Learning fine-grained bimanual manipulation with low-cost hardware},
  author={Zhao, Tony Z and Kumar, Vikash and Levine, Sergey and Finn, Chelsea},
  journal={arXiv preprint arXiv:2304.13705},
  year={2023}
}

@article{yang2026abpolicy,
  title={ABPolicy: Asynchronous B-Spline Flow Policy for Real-Time and Smooth Robotic Manipulation},
  author={Yang, Fan and Jing, Peiguang and Qu, Kaihua and Zhao, Ningyuan and Su, Yuting},
  journal={arXiv preprint arXiv:2602.23901},
  year={2026}
}

@article{li2026smoothvla,
  title={SmoothVLA: Aligning Vision-Language-Action Models with Physical Constraints via Intrinsic Smoothness Optimization},
  author={Li, Jiashun and Shi, Xiaoyu and Xie, Hong and Shang, Mingsheng and Lu, Yun},
  journal={arXiv preprint arXiv:2603.13925},
  year={2026}
}

@article{zhao2025vla,
  title={VLA-RAIL: A Real-Time Asynchronous Inference Linker for VLA Models and Robots},
  author={Zhao, Yongsheng and Zhao, Lei and Cheng, Baoping and Yao, Gongxin and Wen, Xuanzhang and Gao, Han},
  journal={arXiv preprint arXiv:2512.24673},
  year={2025}
}

@article{huang2025notvla,
  title={NoTVLA: Semantics-Preserving Robot Adaptation via Narrative Action Interfaces},
  author={Huang, Zheng and Liu, Mingyu and Lin, Xiaoyi and Zhu, Muzhi and Zhao, Canyu and Du, Zongze and Lin, Ye and Li, Xiaoman and Jia, Yiduo and Zhong, Hao and others},
  journal={arXiv preprint arXiv:2510.03895},
  year={2025}
}

@article{black2026rtc,
  title={Real-time execution of action chunking flow policies},
  author={Black, Kevin and Galliker, Manuel and Levine, Sergey},
  journal={Advances in Neural Information Processing Systems},
  volume={38},
  pages={33383--33407},
  year={2026}
}

@article{pi2025pi05,
  title={$\pi_0.5$:a vision-language-action model with open-world generalization},
  author={Intelligence, Physical and Black, Kevin and Brown, Noah and Darpinian, James and Dhabalia, Karan and Driess, Danny and Esmail, Adnan and Equi, Michael and Finn, Chelsea and Fusai, Niccolo and others},
  journal={arXiv preprint arXiv:2504.16054},
  year={2025}
}

@article{kim2025fine,
  title={Fine-tuning vision-language-action models: Optimizing speed and success},
  author={Kim, Moo Jin and Finn, Chelsea and Liang, Percy},
  journal={arXiv preprint arXiv:2502.19645},
  year={2025}
}

@article{tang2025mind,
  title={Mind to Hand: Purposeful Robotic Control via Embodied Reasoning},
  author={Tang, Peijun and Xie, Shangjin and Sun, Binyan and Huang, Baifu and Luo, Kuncheng and Yang, Haotian and Jin, Weiqi and Wang, Jianan},
  journal={arXiv preprint arXiv:2512.08580},
  year={2025}
}

@misc{pain2021,
      title={High-Resolution Image Synthesis with Latent Diffusion Models}, 
      author={Robin Rombach and Andreas Blattmann and Dominik Lorenz and Patrick Esser and Björn Ommer},
      year={2021},
      eprint={2112.10752},
      archivePrefix={arXiv},
      primaryClass={cs.CV}
}

@article{fei25libero-plus,
    title={LIBERO-Plus: In-depth Robustness Analysis of Vision-Language-Action Models},
    author={Senyu Fei and Siyin Wang and Junhao Shi and Zihao Dai and Jikun Cai and Pengfang Qian and Li Ji and Xinzhe He and Shiduo Zhang and Zhaoye Fei and Jinlan Fu and Jingjing Gong and Xipeng Qiu},
    journal = {arXiv preprint arXiv:2510.13626},
    year={2025},
}

@article{liu2023libero,
  title={LIBERO: Benchmarking Knowledge Transfer for Lifelong Robot Learning},
  author={Liu, Bo and Zhu, Yifeng and Gao, Chongkai and Feng, Yihao and Liu, Qiang and Zhu, Yuke and Stone, Peter},
  journal={arXiv preprint arXiv:2306.03310},
  year={2023}
}

@article{shao2024deepseekmath,
  title={Deepseekmath: Pushing the limits of mathematical reasoning in open language models},
  author={Shao, Zhihong and Wang, Peiyi and Zhu, Qihao and Xu, Runxin and Song, Junxiao and Bi, Xiao and Zhang, Haowei and Zhang, Mingchuan and Li, YK and Wu, Yang and others},
  journal={arXiv preprint arXiv:2402.03300},
  year={2024}
}

@misc{cadene2024lerobot,
    author = {Cadene, Remi and Alibert, Simon and Soare, Alexander and Gallouedec, Quentin and Zouitine, Adil and Palma, Steven and Kooijmans, Pepijn and Aractingi, Michel and Shukor, Mustafa and Aubakirova, Dana and Russi, Martino and Capuano, Francesco and Pascal, Caroline and Choghari, Jade and Moss, Jess and Wolf, Thomas},
    title = {LeRobot: State-of-the-art Machine Learning for Real-World Robotics in Pytorch},
    howpublished = "\url{https://github.com/huggingface/lerobot}",
    year = {2024}
}

@book{kahneman2011thinking,
  title={Thinking, fast and slow},
  author={Kahneman, Daniel},
  year={2011},
  publisher={macmillan}
}

@inproceedings{lugmayr2022repaint,
  title={Repaint: Inpainting using denoising diffusion probabilistic models},
  author={Lugmayr, Andreas and Danelljan, Martin and Romero, Andres and Yu, Fisher and Timofte, Radu and Van Gool, Luc},
  booktitle={Proceedings of the IEEE/CVF conference on computer vision and pattern recognition},
  pages={11461--11471},
  year={2022}
}

@article{pertsch2025fast,
  title={Fast: Efficient action tokenization for vision-language-action models},
  author={Pertsch, Karl and Stachowicz, Kyle and Ichter, Brian and Driess, Danny and Nair, Suraj and Vuong, Quan and Mees, Oier and Finn, Chelsea and Levine, Sergey},
  journal={arXiv preprint arXiv:2501.09747},
  year={2025}
}

@article{intelligence2026pi,
  title={$\pi_0.7$: a Steerable Generalist Robotic Foundation Model with Emergent Capabilities},
  author={Intelligence, Physical and Ai, Bo and Amin, Ali and Aniceto, Raichelle and Balakrishna, Ashwin and Balke, Greg and Black, Kevin and Bokinsky, George and Cao, Shihao and Charbonnier, Thomas and others},
  journal={arXiv preprint arXiv:2604.15483},
  year={2026}
}

@article{bai2025qwen3,
  title={Qwen3-vl technical report},
  author={Bai, Shuai and Cai, Yuxuan and Chen, Ruizhe and Chen, Keqin and Chen, Xionghui and Cheng, Zesen and Deng, Lianghao and Ding, Wei and Gao, Chang and Ge, Chunjiang and others},
  journal={arXiv preprint arXiv:2511.21631},
  year={2025}
}

@article{oquab2023dinov2,
  title={Dinov2: Learning robust visual features without supervision},
  author={Oquab, Maxime and Darcet, Timoth{\'e}e and Moutakanni, Th{\'e}o and Vo, Huy and Szafraniec, Marc and Khalidov, Vasil and Fernandez, Pierre and Haziza, Daniel and Massa, Francisco and El-Nouby, Alaaeldin and others},
  journal={arXiv preprint arXiv:2304.07193},
  year={2023}
}

@article{senin2008dynamic,
  title={Dynamic time warping algorithm review},
  author={Senin, Pavel},
  journal={Information and Computer Science Department University of Hawaii at Manoa Honolulu, USA},
  volume={855},
  number={1-23},
  pages={40},
  year={2008}
}

@article{zhang2020mediapipe,
  title={Mediapipe hands: On-device real-time hand tracking},
  author={Zhang, Fan and Bazarevsky, Valentin and Vakunov, Andrey and Tkachenka, Andrei and Sung, George and Chang, Chuo-Ling and Grundmann, Matthias},
  journal={arXiv preprint arXiv:2006.10214},
  year={2020}
}
\clearpage
\appendix
\section{GESTURE-7: Natural Language-Based End-Effector State Representation}
\label{sec:AppendixA}

We propose \textbf{GESTURE-7} (Gesture Enumerated State Token for Unified Robot End-effector-7D), a natural language-based representation for describing end-effector states. Formally, the state of an end-effector is encoded as a 7-dimensional vector:

$$\mathbf{s} = [x, y, z, r, p, w, g]$$

where $(x, y, z) \in \mathbb{Z}^3$ denote the Cartesian position in centimeter units, $(r, p, w) \in \mathbb{Z}^3$ represent orientation angles (roll, pitch, yaw) in integer degrees, and $g \in \mathcal{G}$ is a discrete gesture token from a predefined vocabulary $\mathcal{G}$ (e.g., {"grasp", "push", "rotate", "release"}). The gesture component uses natural language keywords to indicate the target hand posture, enabling different robot embodiments performing the same action to share identical keywords. This unified representation facilitates cross-embodiment action generalization.

\textbf{Implementation details.} End-effectors exhibit morphological variations; we select representative keypoints for pose representation: the middle finger root for dexterous hands and the gripper center for parallel grippers. To account for mechanical tolerances and hardware imprecisions, we intentionally quantize position predictions to integer centimeters and orientation angles to the nearest degree. This design choice reduces the learning burden on the model and mitigates prediction hallucination.

\begin{figure}[htbp]
    \centering
    \includegraphics[width=0.6\textwidth]{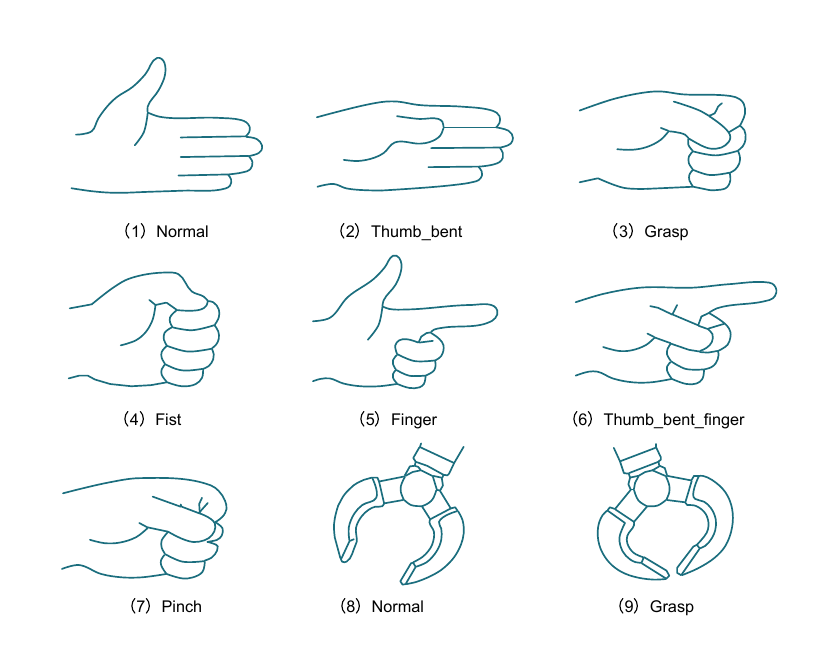}
    \caption{Illustration of common gesture keywords in GESTURE-7 end-effector states. Subfigures (1--7) depict gestures for dexterous hand end-effectors, while subfigures (8--9) depict those for gripper end-effectors.}
    \label{fig:gesture-keywords}
\end{figure}

\textbf{Extensibility.} The gesture vocabulary $\mathcal{G}$ is not fixed; it can be freely extended based on task requirements. \textbf{Figure~\ref{fig:gesture-keywords}} illustrates common gesture keywords for both dexterous hands (subfigures 1--7) and grippers (subfigures 8--9).

\section{System 2 Supervised Fine-Tuning}
\label{sec:AppendixB}
Unlike prior works that incorporate massive visual grounding data or extensive cross-embodiment robot trajectories during pre-training, we observe that large-scale cross-embodiment manipulation data and multimodal perception capabilities yield limited transfer to VLA model training for our target tasks. Moreover, such extensive pre-training may lead to catastrophic forgetting or performance degradation of the original multimodal foundation model. To preserve the model's general capabilities while enabling effective task-specific adaptation, we directly fine-tune on task-specific data using the original Qwen3-VL~\cite{bai2025qwen3} weights as initialization, bypassing additional pre-training stages.

Specifically, we devise five supervised fine-tuning tasks across three categories, all aligned with System~2's two core capabilities: (1)~\textbf{natural language-based state perception}, and (2)~\textbf{GESTURE-7-based trajectory representation and prediction}. The task formulations are detailed below:

\begin{enumerate}
    \item \textbf{Pose Regression:} Given visual observations, predict the current end-effector state in GESTURE-7 format.
    \item \textbf{Keypoint Regression:} From visual observations, regress 2D hand keypoint coordinates~\cite{zhang2020mediapipe}.
    \item \textbf{2D Waypoint Prediction:} From the current frame, predict future trajectory waypoints within the image plane at 2Hz.
    \item \textbf{3D Trajectory Prediction:} Given the current visual frame, predict the robot's end-effector GESTURE-7 trajectory at 2Hz over a future time horizon.
    \item \textbf{Chain-of-Thought (CoT) Generation:} Given the current visual observations, instruction, and end-effector state, generate: (a)~subtask reasoning process, (b)~natural language subtask description, and (c)~future GESTURE-7 trajectory.
\end{enumerate}

These tasks collectively equip System~2 with both precise spatial perception and structured trajectory planning capabilities.

We illustrate the expected CoT output format in Listing~\ref{lst:cot-example}.

\begin{lstlisting}[caption={Example of CoT generation output for Task 5.}, label={lst:cot-example}, breaklines=true, breakatwhitespace=true]
Question: <Multi-view observation> Task: Stack second-order magic cube on top of electrical tape by pinching with two fingers. Current left hand state: (33,21,14,120,74,95,'finger'). Current right hand state: (41,-28,19,5,82,17,'finger'). Output the subtask, reasoning process, and a 3D trajectory in GESTURE-7 format at 2 Hz.

Answer: Subtask: My right hand moves towards the magic cube while my left hand stays still. Reason: My right hand moves to grasp the magic cube for stacking, while my left hand stays still to stabilize the scene and avoid interference. This ensures precise pinching and safe placement atop the electrical tape, maintaining balance and task accuracy. Left_hand_trajectory: [(33,21,14,120,74,95,'finger'), (33,21,14,120,74,95,'finger'), (33,21,14,120,74,95,'finger'), (33,21,14,120,74,95,'finger')]. Right_hand_trajectory: [(40,-27,23,-64,67,-46,'finger'), (40,-25,24,-87,45,-61,'thumb_bent_finger'), (41,-23,19,-104,28,-64,'thumb_bent_finger'), (42,-21,17,-115,16,-61,'thumb_bent_finger')].
\end{lstlisting}

\section{Reward Functions of System 2 Reinforcement Learning}
\label{sec:AppendixC}
Reinforcement learning has demonstrated significant effectiveness in training large language models and vision-language multimodal models. In this work, after System~2 Supervised Fine-Tuning, we employ GRPO~\cite{shao2024deepseekmath} to enhance its trajectory planning capability. During this stage, we further reinforce the five tasks from Stage~1 with specialized reward functions. The reward functions are defined as follows:

\textbf{GESTURE-7 Distance.} Two end-effector states as GESTURE-7 vectors take the form:
$$
\mathbf{s}_i = [x_i, y_i, z_i, r_i, p_i, w_i, g_i]^\top
$$
$$
\mathbf{s}_j = [x_j, y_j, z_j, r_j, p_j, w_j, g_j]^\top
$$

where $(x, y, z) \in \mathbb{Z}^3$ denote the quantized Cartesian position (cm), $(r, p, w) \in \mathbb{Z}^3$ represent orientation angles (roll, pitch, yaw) in integer degrees, and $g \in \mathcal{G}$ is a discrete gesture token from a predefined enumeration set $\mathcal{G}$ (e.g., \texttt{\{grasp, push, rotate, release\}}).

The distance between two GESTURE-7 states, $D(\mathbf{s}_i, \mathbf{s}_j)$, is defined as the weighted sum of three components: \textbf{position distance}, \textbf{orientation distance}, and \textbf{gesture consistency penalty}.

$$
D(\mathbf{s}_i, \mathbf{s}_j) = \lambda_p \cdot d_p(\mathbf{s}_i, \mathbf{s}_j) + \lambda_o \cdot d_o(\mathbf{s}_i, \mathbf{s}_j) + \lambda_g \cdot d_g(\mathbf{s}_i, \mathbf{s}_j)
$$

where $\lambda_p, \lambda_o, \lambda_g \geq 0$ are weighting coefficients (set to $1.0$ each in this work for balanced contributions). Specifically, the position distance $d_p(\mathbf{s}_i, \mathbf{s}_j)$ is defined as the Euclidean distance in Cartesian space; the orientation distance $d_o(\mathbf{s}_i, \mathbf{s}_j)$ is defined as the Euclidean distance in 3D angular space. The gesture distance penalizes mismatches between discrete gesture keywords:

$$
d_g(\mathbf{s}_i, \mathbf{s}_j) =
\begin{cases}
0, & \text{if } g_i = g_j \\[4pt]
1, & \text{if } g_i \neq g_j
\end{cases}
$$

This is a binary (0/1) gesture mismatch indicator.

\textbf{Keypoint Localization Reward.} For the keypoint detection task, the reward function $R_{\text{loc}}$ measures the accuracy of predicted keypoints relative to ground-truth annotations:

$$
R_{\text{loc}} = \frac{1}{K} \sum_{i=1}^{K} \exp\left(-\frac{\|\mathbf{k}_i^{\text{pred}} - \hat{\mathbf{k}}_i\|_2^2}{\sigma^2}\right)
$$

where $K$ is the number of keypoints~\cite{zhang2020mediapipe}, $\mathbf{k}_i^{\text{pred}} \in \mathbb{R}^2$ denotes the predicted 2D coordinates of the $i$-th keypoint, $\hat{\mathbf{k}}_i \in \mathbb{R}^2$ is the corresponding ground-truth annotation, and $\sigma > 0$ is a bandwidth parameter that controls the smoothness of the exponential penalty (larger $\sigma$ yields more forgiving rewards).

\textbf{Waypoints Reward.} Following the approach of Mind to Hand~\cite{tang2025mind}, the waypoints reward combines point-level Euclidean distance with trajectory-level Dynamic Time Warping (DTW) distance~\cite{senin2008dynamic} to evaluate trajectory quality.

\textbf{Trajectory Reward.} End-effector actions are described using the GESTURE-7 format, and the 3D trajectory-level reward is constructed using point-wise GESTURE-7 distance:
$$
R_{\text{traj}} = \frac{1}{K} \sum_{i=1}^{K} (1 - D(\mathbf{s}_i, \hat{\mathbf{s}}_i))
$$
where $\hat{\mathbf{s}}_i$ denotes the ground-truth GESTURE-7 state at trajectory point $i$, and $\mathbf{s}_i$ represents the model's predicted state. Here, $i$ indexes actions sampled at 2Hz. In practice, $K$ typically takes values from $\{2, 3, 4\}$ (corresponding to a temporal horizon of 1--2 seconds). This design reflects our observation that future action predictions become inherently more diverse and uncertain at longer time horizons, making distant trajectory points prone to hallucination.

\textbf{Format Reward.} To ensure structural consistency between inference mode and subtask stages, regex-based matching is employed to enforce a predefined output format. A binary reward (1 or 0) is assigned based on whether the model's output conforms to the predefined format template. This format reward contributes to the total reward with a weight of 0.5.


\end{document}